\documentclass[lettersize,journal]{IEEEtran}
\usepackage{amsmath,amsfonts}
\usepackage{algorithmic}
\usepackage{array}
\usepackage[caption=false,font=normalsize,labelfont=sf,textfont=sf]{subfig}
\usepackage{textcomp}
\usepackage{stfloats}
\usepackage{url}
\usepackage{verbatim}
\usepackage{graphicx}
\def\BibTeX{{\rm B\kern-.05em{\sc i\kern-.025em b}\kern-.08em
    T\kern-.1667em\lower.7ex\hbox{E}\kern-.125emX}}
\usepackage{balance}

\usepackage{orcidlink}
\usepackage{booktabs}
\usepackage{multicol}
\usepackage{multirow}
\usepackage{tabularx}
\usepackage[T1]{fontenc}
\usepackage[utf8]{inputenc}

\usepackage{titlesec}
\titlespacing{\section}{0pt}{5pt}{3pt}
\titlespacing{\subsection}{0pt}{4pt}{2pt}

\usepackage{xcolor}

\begin{document}

\title{
Surrogate-Assisted Genetic Programming with Phenotypic Characterisation in Dynamic Multi-Mode Project Scheduling
}

\author{
Yuan Tian,
% \orcidlink{0000-0001-7435-1557}
\IEEEmembership{Graduate Student Member, IEEE},
Yi Mei,
% \orcidlink{0000-0003-0682-1363}
\IEEEmembership{Senior Member, IEEE},
and Mengjie Zhang,
% \orcidlink{0000-0003-4463-9538}
\IEEEmembership{Fellow, IEEE}%
\thanks{The authors are with the Centre for Data Science and Artificial Intelligence \& School of Engineering and Computer Science, Victoria University of Wellington, New Zealand (e-mail: \{yuan.tian, yi.mei, mengjie.zhang\}@ecs.vuw.ac.nz).}%
\thanks{The code will be made available upon acceptance.}%
% Pre-submission wording
% IEEE Author Center, Submission and Peer Review Policies, Section 3.2:
% https://journals.ieeeauthorcenter.ieee.org/become-an-ieee-journal-author/publishing-ethics/guidelines-and-policies/submission-and-peer-review-policies/
\thanks{This work has been submitted to the IEEE for possible publication. Copyright may be transferred without notice, after which this version may no longer be accessible.}
}

% Add the following code before the \make title command:

\IEEEoverridecommandlockouts

\maketitle

% Add the following code after the \make title command:

\IEEEpubidadjcol

\begin{abstract}
Dynamic multi-mode resource-constrained project scheduling requires decisions to be made under precedence constraints, limited resources, multiple execution modes, and uncertain activity durations. Genetic programming (GP) can automatically evolve heuristic rules for such problems, but its simulation-based fitness evaluation is computationally expensive. This study investigates phenotypic characterisation (PC) in surrogate-assisted GP to evolve higher-quality scheduling heuristics under a fixed budget of full simulation-based fitness evaluations. A key question is how GP individuals should be encoded into phenotypic characterisations to support effective fitness estimation. To answer this question, three PC encoding schemes with different levels of information richness are designed: priority-value encoding, which preserves raw rule outputs; rank encoding, which captures candidate ordering; and binary encoding, which represents final scheduling decisions. These encodings are combined with different distance metrics to measure behavioural similarity between GP individuals.
The experimental results show that binary encoding with Euclidean
distance provides the most effective and robust surrogate guidance. Further analyses show that surrogate estimation accuracy alone does not fully explain the performance differences. The PC representation also determines how effectively phenotypically redundant offspring are removed and how much behavioural diversity is retained after preselection. Ablation experiments further demonstrate that duplicate removal and surrogate preselection provide complementary benefits, with their combination producing the largest improvement. These findings highlight that effective surrogate-assisted GP depends not only on identifying promising offspring, but also on controlling redundancy and preserving useful diversity during evolutionary search.
\end{abstract}


\section{Introduction}
% What's DMRPCPSP
\IEEEPARstart{P}{ROJECT} scheduling plays an important role in many real-world applications, such as supply chain management \cite{asadujjamanSupplyChainIntegrated2024}, manufacturing \cite{rahmanEnergyefficientProjectScheduling2022}, and construction \cite{liuBiobjectiveOptimizationResourceconstrained2023}, where a set of interrelated activities must be completed under precedence and resource constraints. In practice, project environments are often dynamic and uncertain, as activity durations may deviate from their initial estimates due to technical risks, resource availability, or external disruptions. Therefore, instead of generating a fixed schedule before project execution, dynamic project scheduling requires scheduling decisions to be made or revised as new information becomes available during execution. This challenge becomes more complex in the dynamic multi-mode project scheduling problem, where each activity can be executed in one of several modes, representing different trade-offs between duration and resource requirement. An effective scheduling method needs to make two closely related decisions: selecting which activities should be scheduled at each decision point, and choosing an appropriate execution mode for each selected activity.

% Methods to solve DMRCPSP.
For dynamic project scheduling problems, heuristic rules \cite{liEfficientPriorityRules2026} are commonly used because they can make real-time decisions based on the current project state. However, designing effective heuristic rules manually is challenging, as good scheduling decisions often require a careful balance between multiple and sometimes conflicting factors. A rule that performs well in one project scenario may not generalise well to other scenarios, especially when project structures, resource constraints, and uncertainty levels change. Genetic programming (GP) \cite{zhang_survey_2023} provides a promising way to address this limitation by automatically evolving heuristic rules from combinations of scheduling features and mathematical operators. Instead of relying on manually designed priority rules, GP can search for effective rule structures that adapt to different decision situations and capture complex interactions.

% Introduction of surrogate model
Despite its potential, GP usually requires a large number of fitness evaluations to evolve effective heuristic rules. In dynamic multi-mode project scheduling, evaluating a GP individual is particularly expensive because each rule needs to be tested through simulation-based scheduling over multiple project instances. This computational cost becomes more significant for large-scale problems, where many activities, execution modes, and resource constraints need to be considered repeatedly during the evolutionary process.

Surrogate-assisted GP \cite{hildebrandt_using_2015, nguyen_surrogate-assisted_2017} has been used to reduce this computational burden by replacing some expensive fitness evaluations with cheaper fitness estimations. Existing surrogate models in GP can be broadly divided into simplified simulation-based surrogates and phenotypic characterisation (PC)-based surrogates. Simplified simulation-based surrogates estimate fitness by using a less expensive version of the original evaluation procedure, such as fewer simulation samples or simplified problem instances. In contrast, PC-based surrogates estimate fitness by measuring the behavioural similarity between GP individuals. A key step in this type of surrogate is to map each GP individual into a numerical vector that characterises its behaviour, so that a distance metric can be used to find the nearest evaluated individuals in the surrogate database.
PC-based surrogate GP has shown promising performance in domains such as (flexible) job shop scheduling \cite{hildebrandt_using_2015,zhang_surrogate-assisted_2021,zhu_phenotype_2025} and single-mode project scheduling \cite{chen_surrogate-assisted_2023,chenSurrogateassistedGeneticProgramming2025}. However, PC design is highly problem-dependent. Existing PC encodings developed for these problems cannot be directly applied to dynamic multi-mode project scheduling, where scheduling decisions involve both activity selection and mode selection under changing project states. In addition, group-selection-based GP \cite{tian_genetic_2025,tian_scalable_2025} requires the behaviour of activity group selection to be characterised, rather than only the priority ordering of individual activities. Therefore, applying PC-based surrogate GP to DMRCPSP requires problem-specific phenotypic characterisations that can represent the scheduling behaviour of multi-tree GP individuals.

% Research gap
However, designing a problem-specific PC is only the first step. Distance measurement between individuals should be considered for most surrogate models, as different distance metrics may produce different neighbourhood structures in the surrogate database, which can directly affect the performance of fitness estimation and the quality of offspring preselection. Despite its importance, the impact of PC encodings and distance metrics on surrogate-assisted GP for DMRCPSP has not been systematically investigated. It remains unclear how these design choices affect surrogate performance and final scheduling performance.

% Motivation
Motivated by these limitations, this paper focuses on PC design for surrogate-assisted GP in dynamic multi-mode project scheduling. The goal of this paper is to determine how GP individuals should be characterised and compared so that the surrogate model can reliably identify promising offspring. This provides the basis for analysing the role of behavioural representation in surrogate-assisted GP for DMRCPSP.
% Goal of this paper.
The main contributions of this paper are summarised as follows.
\begin{enumerate}
\item This paper develops PC-based surrogate-assisted GP for dynamic multi-mode project scheduling under a fixed budget of full fitness evaluations. The approach generates an enlarged intermediate-offspring pool and combines PC-based duplicate removal with surrogate preselection to make more effective use of this budget.

\item This paper designs problem-specific PCs for both activity-mode pair selection and activity group selection, and investigates three encoding schemes with different levels of behavioural abstraction. Priority-value encoding retains the original priority values produced by GP rules, rank encoding represents their relative ordering, and binary encoding records only the final selection decisions. These encodings are combined with different distance metrics for PC-based similarity evaluation.

\item This paper conducts extensive experiments and further analyses to evaluate optimisation performance, evaluation efficiency, computational overhead, surrogate estimation and preselection quality, duplicate-removal capacity, and offspring diversity. An ablation study further examines the individual and combined effects of duplicate removal and surrogate preselection. The results show that effective surrogate-assisted GP depends not only on identifying promising offspring, but also on reducing phenotypic redundancy and maintaining useful offspring diversity.

\end{enumerate}

\section{Background and Related Work}
\subsection{Problem Description}
% Problem Statement
The dynamic multi-mode resource-constrained project scheduling problem (DMRCPSP) considered in this study involves a project consisting of a set of activities $A=\{0,1,...,n,n+1\}$, where activities $0$ and $n+1$ denote the dummy start and end activities, respectively. The activities are subject to precedence relations represented by a directed acyclic graph $G=(A,E)$, where $(i,j)\in E$ indicates that activity $j$ cannot start until activity $i$ has been completed.
The project contains a set of renewable resources $R$, where each resource type $r\in R$ has a limited per-time-unit capacity $K_r$. These resources are termed renewable because up to $K_r$ units are available at every time unit throughout the project horizon, and the occupied units are released once ongoing activities are completed.
Each non-dummy activity $i\in A \setminus\{0,n+1\}$ can be executed in one mode selected from a predefined mode set $M_i$. Each mode $m \in M_i$ specifies both the expected duration $\hat{d}_{i,m}$ of the activity and its resource requirement $k_{i,m,r}$ for each resource type $r \in R$. To model duration uncertainty in the dynamic environment, each activity-mode pair $(i,m)$ is associated with an optimistic duration $d^{\text{min}}_{i,m}$ and a pessimistic duration $d^{\text{max}}_{i,m}$. The realised duration $d_{i,m}$ is sampled from $[d^{\text{min}}_{i,m}, d^{\text{max}}_{i,m}]$ and remains unknown until the activity starts.

% The problem can be formulated as:
% \begin{equation}
%     \text{min} C_{\text{max}}
% \end{equation}
% subject to

% \begin{equation}
% \label{eq:precedence_constraint}
%     s_j \geq s_i+d_{i,m}, \forall (i,j) \in E
% \end{equation}

% \begin{equation}
% \label{eq:resource_onstraint}
%     \sum_{i \in S_t}{k_{i,m,r}} \leq K_r, \forall r \in R, \forall t
% \end{equation}

% \begin{equation}
% \label{eq:actual_duration_sample}
%     d_{i,m} \in [d^{\text{min}}_{i,m}, d^{\text{max}}_{i,m}]
% \end{equation}
% Here, $s_i$ denotes the start time of activity $i$, $S_t$ denotes the ongong activity set. Constraint (\ref{eq:precedence_constraint}), Constraint (\ref{eq:resource_onstraint}), and Constraint (\ref{eq:actual_duration_sample}) correspond to precedence feasibility, renewable resource feasibility, and duration uncertainty, respectively.

% Place the example data early so Table I precedes the schedule figure.
\begin{figure}[t]
    \centering
    \includegraphics[width=.5\linewidth]{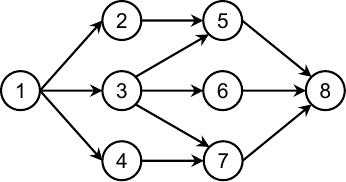}
    \caption{Precedence graph of the DMRCPSP example.}
    \label{fig:precedence_relation_example}
\end{figure}

\begin{table}[!t]
    \caption{Activity-mode information of the example, including duration estimates and renewable resource requirements.}
    \label{tab:project_data_example}
    \centering
    \renewcommand{\arraystretch}{0.9}
    \begin{tabular}{ccccccc}
        \toprule
        Activity $i$ &  Mode  & $\hat{d}_{i,m}$ & $d^{\text{min}}_{i,m}$  & $ d^{\text{max}}_{i,m}$ & R1 & R2\\
        \midrule
        1 & 1 & 0 & 0 & 0  & 0  & 0\\
        2 & 1 & 5 & 3 & 7  & 10 & 9\\
          & 2 & 7 & 5 & 8  & 6  & 4\\
        3 & 1 & 4 & 3 & 7  & 7  & 0\\
          & 2 & 6 & 5 & 7  & 5  & 0\\
        4 & 1 & 4 & 3 & 6  & 9  & 6\\
          & 2 & 6 & 5 & 8 & 5  & 5\\
        5 & 1 & 4 & 2 & 5  & 7  & 8\\
          & 2 & 6 & 4 & 8  & 4  & 4\\
        6 & 1 & 3 & 2 & 5  & 9  & 10\\
          & 2 & 6 & 4 & 7  & 6  & 7\\
        7 & 1 & 3 & 2 & 5  & 9  & 8\\
          & 2 & 5 & 4 & 7  & 6  & 4\\
        8 & 1 & 0 & 0 & 0 &  0  & 0\\
        \midrule
        \multicolumn{4}{l}{Resource availability} & & 12 & 10\\
        \bottomrule
    \end{tabular}
\end{table}

During project execution, the total resource demand for each resource type from all ongoing activities cannot exceed its available capacity.
Scheduling decisions are triggered whenever one or more ongoing activities are completed or the resource state is updated. At each decision point, an eligible activity set $\varepsilon_t$ is formed, containing all activities whose predecessors have been completed and are ready for execution. The scheduler then dynamically selects one or more activities from $\varepsilon_t$ and assigns an execution mode to each selected activity according to the current project state.
The objective is to minimise the makespan of the project $C_\text{max}$ while satisfying all precedence and resource constraints throughout project execution.

An example in Fig.~\ref{fig:precedence_relation_example} and Table~\ref{tab:project_data_example} contains six non-dummy activities, each with two modes, and dummy start and end activities numbered 1 and 8. The table lists mode durations and resource requirements; resources R1 and R2 have capacities of 12 and 10 units, respectively.

\begin{figure}[t]
    \centering
    \includegraphics[width=0.8\linewidth]{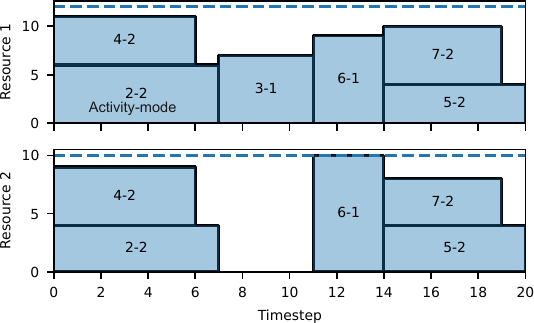}
    \caption{A feasible schedule for the illustrative example.}
    \label{fig:example_solution}
\end{figure}
The feasible schedule in Fig.~\ref{fig:example_solution} has a makespan of 20 when realised durations equal their expected values. Rectangle widths and heights represent activity durations and resource consumption, respectively; the dashed lines indicate resource capacities.

% Fig. \ref{fig:example_solution} shows a feasible schedule under the assumption that the expected durations are equal to the realised actual durations. The schedule is visualised in terms of resource usage profiles, illustrating how activities occupy the two resource types over time. The horizontal axis represents the time steps, while the vertical axis shows the amount of resource units in use. The blue horizontal dashed lines indicate the resource capacities.
% For each activity-mode pair, the duration of execution and the amount of resource consumption are represented by the width and height of a rectangle, respectively.
% Rectangles corresponding to simultaneously executing activities are stacked vertically, and feasibility is maintained as long as the total height does not exceed the horizontal capacity line.
% Since an activity may not require all types of resources, some activities may not appear in every resource profile. For example, the rectangle corresponding to Activity 3 does not appear in the R2 profile because this activity does not consume Resource 2. Such temporary resource idleness is unavoidable, as subsequent Activities 5, 6, and 7 cannot start until Activity 3 has been completed.
% The final project makespan is determined by the completion time of the latest finishing non-dummy activity, which is Activity 5 in this example, resulting in a makespan of 20.
% The final project makespan is 20.

\subsection{Genetic Programming for Project Scheduling}
Genetic Programming (GP) is a hyper-heuristic approach that automatically evolves scheduling heuristics for complex scheduling problems.
GP \cite{zhang_survey_2023} has been successfully applied to a range of dynamic and uncertain optimisation problems that require real-time decision making, such as job shop scheduling (JSP) \cite{zhang_survey_2023}, arc routing \cite{wang_explainable_2023}, and air traffic control \cite{guo_genetic_2024}.
% Among these, JSP is one of the most extensively studied problems in GP-base hyper-heuristics.
To further enhance the applicability, performance, and interpretability of GP, various machine learning techniques have been incorporated, including feature selection \cite{shadyFeatureSelectionApproach2023,zhang_evolving_2021}, ensemble learning \cite{xuGeneticProgrammingDynamic2023}, multi-task learning \cite{zhangMultitaskGeneticProgrammingBased2022}, and multi-objective learning \cite{zhangMultitaskMultiobjectiveGenetic2023}.

% GP for project scheduling.
The application of GP to the classical single-mode static RCPSP can be traced back to early work in \cite{frankolaEvolutionaryAlgorithmsResource2008}. Since then, a variety of aspects have been explored, including representation design \cite{chand_use_2018}, fitness function \cite{dumicEvolvingPriorityRules2018}, and rollout-based approaches \cite{dumicEvolvingPriorityRules2018, umic_using_2022} for GP in RCPSP. Given the diverse variants of RCPSP, GP has also been extended to more complex settings, such as multi-project scheduling \cite{chaoGeneticProgrammingHyperheuristic2026,chen_surrogate-assisted_2023}, spatially constrained problems \cite{liNovelHyperheuristicBased2025}, and dynamic scenarios involving events like resource disruptions \cite{chandEvolvingHeuristicsResource2019} and activity insertions \cite{chenFilteringGeneticProgramming2022,chenTwostageGeneticProgramming2022}. Due to the common characteristics of GP in scheduling—where individuals typically act as priority functions—many advanced techniques developed in other scheduling domains, such as feature selection \cite{chenFilteringGeneticProgramming2022} and ensemble learning \cite{dumic_ensembles_2021, chenHyperheuristicBasedEnsemble2021}, have also been adapted for RCPSP.

% GP for DMRCPSP.
Compared to other RCPSP variants, dynamic multi-mode RCPSP (DMRCPSP) introduces additional complexity by allowing each activity to be executed in multiple modes. This significantly increases the diversity of possible schedules and makes decision making more challenging. Instead of selecting only activities from the eligible set, both the activity and its execution mode must be determined. As a result, research on GP for DMRCPSP has mainly focused on designing more effective decision-making strategies and leveraging GP to evolve corresponding heuristic rules.
For example, the study in \cite{tian_learning_2024} investigated different decision sequences for selecting activities and modes within the eligible set, including selecting activities first and then assigning modes, determining modes before selecting activities, and treating activity–mode combinations as independent decision units. More recently, activity group selection strategies \cite{tian_genetic_2025,tian_scalable_2025} have been proposed, where multiple activity–mode pairs are combined into feasible groups that maximise resource utilisation, and a group is selected based on priority evaluation. Although such approaches have demonstrated strong performance, they typically incur higher computational costs due to the need to evaluate multiple candidate groups at each decision point. This issue is particularly critical in GP, where large numbers of individuals must be evaluated through simulations.
Therefore, evolving effective scheduling heuristics within a limited budget of full fitness evaluations remains an important research challenge for GP in DMRCPSP.

% \subsection{Surrogate Models in Genetic Programming for Evolving Scheduling Heuristics}

\subsection{Surrogate-Assisted GP for Evolving Scheduling Heuristics}
Surrogate models provide inexpensive approximations of computationally expensive fitness evaluations~\cite{jin_surrogate-assisted_2011}. In GP for evolving scheduling heuristics, these models estimate the fitness of candidate rules to guide evolutionary search. Existing approaches can be broadly divided into simplified simulation-based approaches and phenotypic characterisation (PC)-based approaches.
% half shop
\subsubsection{Simplified Simulation-Based Approaches}
Simplified model-based surrogate approaches aim to reduce the computational cost of fitness evaluation by replacing the original simulation with a simplified version, typically constructed using smaller-scale problem instances. In this framework, GP individuals are evaluated on reduced simulations that approximate the original problem while significantly lowering computational overhead.
For example, \cite{nguyen_surrogate-assisted_2017} investigated simplified models in the context of job shop scheduling by analysing the rank correlation between simplified and original simulations, as well as the performance of GP under different simplification levels. Their results showed that a half-shop configuration, using half of the jobs and machines from the original problem, provides a good trade-off between accuracy and efficiency. This idea has also been extended to other domains, including electric vehicle routing \cite{gil-galaGeneticProgrammingSurrogate2025}, air traffic flow management \cite{guo_genetic_2024}, and RCPSP \cite{luoAutomatedDesignPriority2023}.
Different strategies have been proposed to utilise simplified models. In \cite{gil-galaGeneticProgrammingSurrogate2025}, a two-level evaluation framework was introduced, where all offspring are first evaluated on a smaller training set, and only the top-performing individuals are further evaluated on the full training set. In \cite{guo_genetic_2024}, multiple simplified models with different fidelity levels were constructed by adjusting the simulation timespan, and their accuracy and computational overhead were monitored for surrogate management. In \cite{luoAutomatedDesignPriority2023}, the configuration of simplified models for static single-mode RCPSP was systematically studied, analysing the effects of instance scale, schedule generation schemes, and resource scarcity on evaluation accuracy.
These studies demonstrate that simplified models can effectively reduce evaluation cost while preserving relative performance to a certain extent. However, their effectiveness is often highly dependent on the problem characteristics and the choice of instance configurations. Changes in problem settings or training instance distributions may lead to different conclusions. Moreover, identifying appropriate simplification configurations typically requires extensive empirical testing, which can itself be computationally expensive.
% PC in JSP
\subsubsection{Phenotypic Characterisation-Based Approaches}
% Why cannot directly in this problem.
Phenotypic characterisation (PC)-based surrogate approaches are built on the assumption that GP individuals with similar decision-making behaviours are likely to have similar fitness values. A key challenge in GP for scheduling heuristics is therefore how to quantify the behavioural characteristics of GP individuals in a meaningful and computationally efficient way.
Early work in \cite{hildebrandt_using_2015} proposed one of the first PC schemes for GP applied to JSP. In this approach, GP individuals were evaluated on a set of decision situations extracted from simulation, and their behaviours were characterised by observing how they selected among eligible jobs. The candidate jobs in each decision situation were first ranked and indexed using a reference rule. The decision made by a candidate GP rule was then encoded by the index of the selected job under the reference-rule ranking.
By recording the PC vectors and true fitness values of previously evaluated individuals, a surrogate database can be constructed. For a new unevaluated GP individual, its PC vector is first generated and then compared with those stored in the database. Based on the most similar PC vectors, for example using a k-nearest neighbour model, its fitness can be estimated.

Compared with simplified model-based surrogates, PC-based approaches are typically more lightweight, as they do not require running complete simulations. This can substantially reduce computational cost. In addition, PC vectors can be used for other purposes, such as clustering individuals \cite{zhu_phenotype_2025}, measuring subtree importance \cite{zhangCorrelationCoefficientBasedRecombinative2021}, and diversity selection \cite{xuSemanticGeneticProgramming2024,xu_quality_2025,xuGeneticProgrammingAdvanced2026}. Subsequent studies further extended the original PC framework by considering different types of decision situations, addressing the issue that fitness values obtained from different instance sets across generations may not be directly comparable \cite{zhangInstanceRotationBasedSurrogateGenetic2023}, and combining phenotypic and genotypic information in surrogate modelling \cite{zhu_phenotype_2025,tan_pgu-sgp_2025}.

Despite these advantages, relatively little attention has been paid to the design of PC encoding schemes and the choice of distance metrics between PC vectors. This is an important research issue for several reasons. First, even under the same problem and the same decision situations, different encoding schemes may capture different aspects of behaviour and therefore induce different surrogate landscapes. The widely used reference-rule-based rank encoding is efficient, but it only records the final choice made by a GP rule and ignores how the rule ranks the remaining candidates in the eligible set. Moreover, this encoding depends on the reference rule, meaning that the resulting PC-vector distance reflects an indirect relationship through the reference rule, rather than the direct behavioural distance between two GP individuals. To address this limitation, \cite{chenNeuralNetworkSurrogate2025} proposed using priority values directly as PC vectors and employed a neural network to build the surrogate model.

Second, even for the same PC encoding scheme, different distance metrics may affect surrogate accuracy by altering the neighbourhood structure among PC vectors. Most existing studies use Euclidean distance because of its simplicity. In \cite{chen_surrogate-assisted_2023,chenSurrogateassistedGeneticProgramming2025}\, a distance measure based on the number of swaps between the reference-rule ranking and the ranking induced by the rule under comparison was proposed, but its difference from Euclidean distance was not systematically investigated.

Third, although the reference-rule-based rank encoding has been widely applied in problems such as JSP \cite{hildebrandt_using_2015}, terminal truck scheduling \cite{tan_pgu-sgp_2025}, and RCPSP \cite{chen_surrogate-assisted_2023,chenSurrogateassistedGeneticProgramming2025}, it is not directly suitable for the decision-making scenario considered in DMRCPSP. Existing PC schemes are mainly designed for one-of-many decisions, where a single job or activity is selected from the eligible set. In contrast, the activity ordering procedure in DMRCPSP requires selecting a subset of activity-mode pairs from the eligible set. A reference rule can only represent the ranking of individual activity-mode pairs, but cannot directly encode the selection of a subset. Even without using a reference rule, directly enumerating possible subsets for index-based encoding would lead to a combinatorial explosion. For example, with 10 activities and three modes per activity, the number of possible subsets would be approximately \(4^{10} \approx 10^6\). To better suit this setting, \cite{tian_surrogate-assisted_2026} proposed a preliminary PC design that directly encodes the rankings of activity–mode pairs induced by GP rules. However, this work only provided an initial exploration and did not investigate the differences among alternative encoding schemes.

Motivated by these research gaps, this study investigates different PC encoding schemes and distance metrics in the context of GP for DMRCPSP, and examines their effects on surrogate modelling performance and the effectiveness of surrogate-assisted GP.
% Limitation of current study.

\section{Methodology}
\subsection{Overall Framework}
\label{section:overall_framework}
\begin{figure}[!t]
    \centering
    \includegraphics[width=\linewidth]{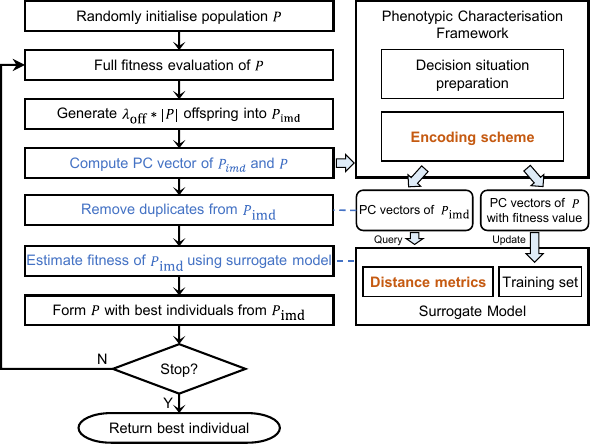}
    \caption{Overall framework of the surrogate-assisted GP algorithm.}
    \label{fig:surrogate_assisted_GP_flowchart}
\end{figure}

Figure~\ref{fig:surrogate_assisted_GP_flowchart} summarises the PC-based surrogate-assisted GP framework adopted in this study. Following~\cite{hildebrandt_using_2015}, the framework uses phenotypic characterisation for surrogate fitness estimation and offspring preselection. This study applies the framework to DMRCPSP and investigates problem-specific PC encoding schemes and distance metrics. The interactions among GP, PC, and the surrogate model are highlighted in blue, while the components investigated in this study are shown in orange.

\begin{figure}[!t]
    \centering
    \includegraphics[width=0.7\linewidth]{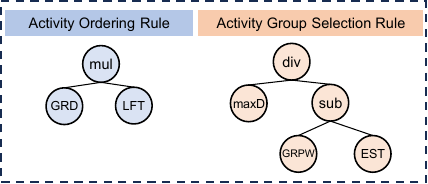}
    \caption{An example of a GP individual consisting of two tree-based heuristic rules for DMRCPSP.}
    \label{fig:GP_individual_example}
\end{figure}

For the GP component, this study adopts the multi-tree representation and knee-point-guided activity group selection mechanism in~\cite{tian_scalable_2025}. Each GP individual consists of two tree-based heuristic rules, as illustrated in Fig.~\ref{fig:GP_individual_example}: an activity ordering rule and an activity group selection rule. The two rules serve different roles in schedule generation. The activity ordering rule prioritises eligible activity--mode pairs to support the construction of promising subsets, whereas the activity group selection rule evaluates the resulting feasible activity groups and selects one for execution. The scheduling procedure is described in Section~\ref{section:scheduling_procedure} because its decisions provide the behavioural information used for PC construction.

Prior to evolution, the decision situations used for PC construction are prepared, and the PC encoding scheme and distance metric are specified. Their designs are described in Sections~\ref{section:PC_framework} and~\ref{section:surrogate_model}, respectively. The algorithm then initialises a population $P$ and evaluates its individuals through scheduling simulations on the training instances.

At each generation, an intermediate offspring population $P_{\text{imd}}$ of size $\lambda_{\text{off}}\times|P|$ is generated. PC vectors are constructed for both the evaluated population and the intermediate offspring. The PC vectors and true fitness values of the evaluated individuals are used to update the surrogate database. Duplicate intermediate offspring with identical PC vectors are then removed, after which the surrogate model estimates the fitness of the remaining offspring. Based on the estimated fitness values, the best $|P|$ individuals are selected from $P_{\text{imd}}$ to form the population $P$ for the next generation. This selection constitutes surrogate-assisted preselection. The resulting population is subsequently evaluated through scheduling simulations. This process continues until the termination criterion is met, and the best evaluated individual is returned.

\subsection{Heuristic-Guided Scheduling Procedure}
\label{section:scheduling_procedure}

\begin{figure}[t]
    \centering
    \includegraphics[width=\linewidth]{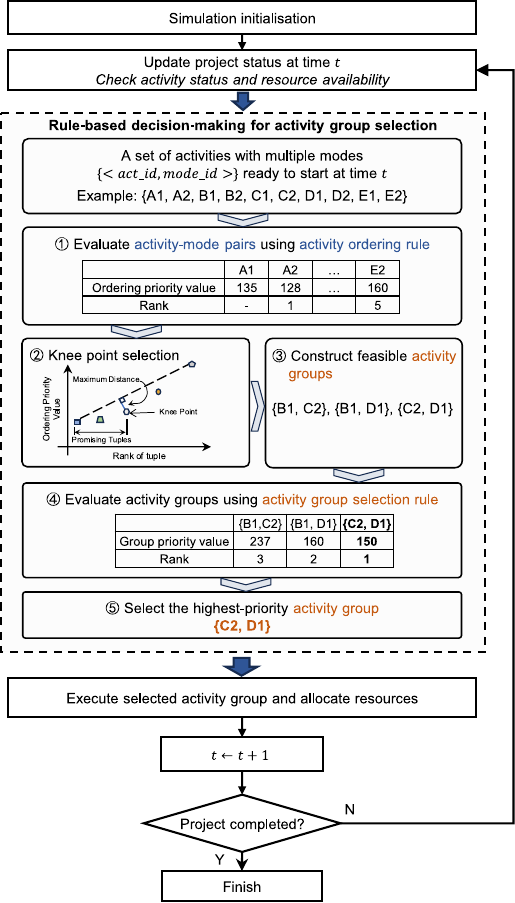}
    \caption{Flowchart of the heuristic-guided scheduling procedure for DMRCPSP. The dashed region represents the activity group selection process guided by the two tree-based heuristic rules.}
    \label{fig:decision_making_flowchart}
\end{figure}

This study uses the scalable knee-point-guided activity group selection procedure in~\cite{tian_scalable_2025} as the underlying schedule-generation mechanism. The procedure is summarised here for completeness and because its activity ordering and activity group selection decisions form the basis of the phenotypic characterisations developed in this study.

To evaluate a GP individual, its two tree-based heuristic rules are embedded into a discrete-event simulation of project execution. As shown in Fig.~\ref{fig:decision_making_flowchart}, the simulation starts by loading a project instance and updating the project status at each decision time point $t$, including activity completion status and resource availability. Based on the current project state, an eligible set of activity--mode pairs is identified.

The dashed region in Fig.~\ref{fig:decision_making_flowchart} represents the heuristic-guided activity group selection process. First, the activity ordering rule assigns a priority value to each eligible activity--mode pair. Promising pairs are then identified using the knee-point selection strategy~\cite{zhang_evolving_2021}. Based on these selected pairs, feasible activity groups are constructed such that all activities within a group can be executed simultaneously without violating the resource constraints. The activity group selection rule subsequently evaluates these candidate groups, and the group with the lowest priority value is selected for execution.
After the selected activity group has been scheduled and the corresponding resources have been allocated, the simulation advances to the next decision time point. This process is repeated until all activities have been completed, after which the resulting schedule and project makespan are returned.

\subsection{Phenotypic Characterisation}
\label{section:PC_framework}
\subsubsection{Decision Situation Preparation}
The construction of PC vectors relies on a predefined set of decision situations. By grounding the PC representation in these decision situations, the behavioural characteristics of GP individuals can be evaluated under consistent and representative decision-making contexts.

\begin{figure}
    \centering
    \includegraphics[width=\linewidth]{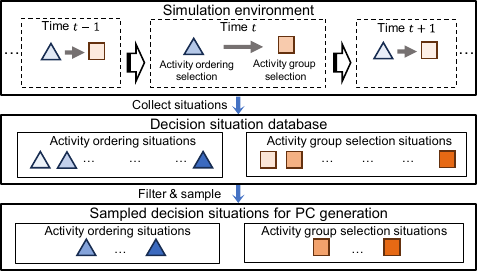}
    \caption{Decision situation collection process.}
    \label{fig:decision_situations_collection}
\end{figure}

The overall procedure for collecting these situations is illustrated in Fig. \ref{fig:decision_situations_collection}. These decision situations are extracted from the simulation environment and reflect the actual decision-making contexts encountered during the scheduling process.
At each decision time step $t$, two types of decisions are involved, namely activity ordering and activity group selection. During the simulation, all encountered decision situations are recorded. Each situation captures the complete state information required for decision making, including the set of eligible activity–mode pairs, resource availability, and the status of completed and ongoing activities.
After collection, the decision situations are categorised according to their decision types. A subset of representative situations is then selected based on predefined filtering criteria and the required number of situations. These selected decision situations are subsequently used to construct the PC vectors.

\subsubsection{Encoding Scheme Design}

To construct the PC vectors, it is necessary to define how the behavioural responses of GP individuals are encoded for each decision situation. In this paper, three encoding schemes are considered and presented in descending order of information richness, namely (priority-)value-based encoding, rank-based encoding, and binary encoding. Each encoding scheme maps a GP individual's response to a decision situation into a decision vector, which will later be aggregated to form the PC vector.

\textbf{(Priority-)value-based encoding} preserves the most detailed information of the GP individual's decision process. In activity ordering situations, it records the priority value assigned to each activity–mode pair. In activity group selection situations, it records the priority value assigned to each feasible activity group. This encoding is based on the assumption that two behaviourally similar individuals will produce similar priority values for candidates across different decision situations.

\textbf{Rank-based encoding} records the ranking of candidates induced by the priority values produced by the GP individual. In activity ordering situations, the activity–mode pairs are ranked in ascending order of priority value. In activity group selection situations, the feasible activity groups are ranked in ascending order of priority value. This encoding assumes that two similar individuals will generate similar rankings across decision situations, even if their raw priority values are not identical.

\textbf{Binary encoding} records only the final choices made by the GP individual in each decision situation. In activity ordering situations, the selected activity–mode pairs are recorded. In activity group selection situations, the activity–mode pairs in the selected group are recorded. Selected items are represented by 1, while unselected items are represented by 0. Compared with the other two schemes, this encoding is based on a looser assumption: individuals are regarded as similar as long as they make similar final choices, regardless of the underlying priority values or intermediate rankings.

\begin{figure}
    \centering
    \includegraphics[width=0.9\linewidth]{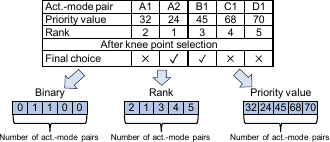}
    \caption{Example of activity-mode subset selection encoding using various schemes.}
    \label{fig:activity_subset_selection_encoding_scheme}
\end{figure}

\begin{figure}
    \centering
    \includegraphics[width=0.85\linewidth]{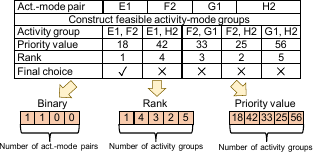}
    \caption{Example of activity group selection encoding using various schemes.}
    \label{fig:activity_group_selection_encoding_scheme}
\end{figure}

Figures \ref{fig:activity_subset_selection_encoding_scheme} and \ref{fig:activity_group_selection_encoding_scheme} provide examples of the three encoding schemes for activity ordering and activity group selection situations, respectively. In activity ordering, the lengths of the decision vectors under all three encoding schemes are equal to the number of activity–mode pairs. In activity group selection, however, the length of the binary decision vector is equal to the number of activity–mode pairs, whereas the lengths of the priority-value-based and rank-based decision vectors are equal to the number of feasible activity groups.

\subsubsection{PC Vector Construction}
\begin{figure}[!t]
    \centering
    \includegraphics[width=\linewidth]{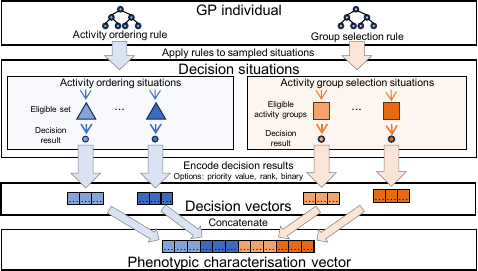}
    \caption{Phenotypic characterisation vector generation.}
    \label{fig:PC_generation}
\end{figure}

The PC vector of a GP individual is constructed in a two-step manner, and the overall process of PC vector construction is illustrated in Fig. \ref{fig:PC_generation}. For each decision situation, the GP individual applies its heuristic rules (i.e., activity ordering rule or activity group selection rule) to make decisions. The results are then encoded into a decision vector according to the selected encoding scheme. The PC vector is subsequently obtained by concatenating the decision vectors across all selected decision situations.

\subsection{Surrogate Model}
\label{section:surrogate_model}
\subsubsection{Nearest-Neighbour-Based Fitness Estimation}

% {\color{blue} The choice of surrogate model is not the primary focus of this study. Following previous studies in surrogate-assisted GP, where 1-nearest neighbour (1-NN) models combined with phenotypic characterisation (PC) have been successfully employed for fitness approximation~\cite{hildebrandt_using_2015,zhang_surrogate-assisted_2021,chenSurrogateassistedGeneticProgramming2025}, this study adopts 1-NN as the surrogate model and focuses instead on how GP individuals should be represented and compared in the PC space.}

% Given a query individual represented by its PC vector, the 1-NN model identifies the most similar individual in the training data according to the selected distance metric and uses the fitness of this nearest neighbour as the estimated fitness of the query individual. This model is particularly suitable for PC-based surrogate modelling because PC provides an explicit behavioural representation of GP individuals, allowing their similarity to be naturally quantified through distance measures. Moreover, as a lazy learning method, 1-NN requires no explicit model-training phase, which makes it computationally inexpensive and straightforward to update as newly evaluated individuals become available during evolution.

Following~\cite{hildebrandt_using_2015,zhang_surrogate-assisted_2021,chenSurrogateassistedGeneticProgramming2025}, we use a 1-nearest neighbour (1-NN) surrogate to focus the comparison on PC encodings and distance metrics. A query individual's fitness is estimated using the evaluated fitness of its nearest neighbour in PC space under the selected distance metric. As a lazy learning method, 1-NN requires no explicit training and can incorporate newly evaluated individuals by updating its database.

\subsubsection{Distance Metrics}
A set of distance metrics is considered, including conventional distance-based functions and correlation-derived distance metrics. Table \ref{tab:distance_functions_encoding} summarises the compatibility between encoding schemes and distance metrics.
% Let $z^{(s)}_i$ and $z^{(s)}_j$ denote the decision vectors in situation $s$ of two individuals. The overall distance between two individuals is computed in a situation-wise manner as
% {
% \setlength{\abovedisplayskip}{4pt}
% \setlength{\belowdisplayskip}{4pt}
% \begin{equation}
%     D(i,j) = \sum^{S}_{s=1}{d(z_{i}^{(s)}, z_{j}^{(s)})}
% \end{equation}
% }
% where $d(\cdot,\cdot)$ is defined according to the selected similarity metric.

\textbf{Euclidean distance} is used as a general-purpose metric and can be applied to all encoding schemes. For priority-value-based encoding, min-max normalisation is first applied to the priority values within each decision situation to prevent large priority-value ranges from dominating the distance. For each encoding scheme, the resulting decision vectors from all situations are concatenated into a complete PC vector for each individual, and the Euclidean distance is computed directly between these PC vectors.

\textbf{Kendall tau distance}~\cite{kendall1938rank} is used for rank-based encoding to capture differences in pairwise ordering between candidates. For a decision situation $s$, the Kendall tau distance between two decision vectors is computed as the number of discordant candidate pairs:
{
\setlength{\abovedisplayskip}{4pt}
\setlength{\belowdisplayskip}{4pt}
\begin{equation}
d_{\tau}(z_{i}^{(s)}, z_{j}^{(s)})=\sum_{p<q}{\mathbb{I}[(z_{i,p}^{(s)}-z_{i,q}^{(s)})(z_{j,p}^{(s)}-z_{j,q}^{(s)})<0]}
\end{equation}
}
where $\mathbb{I}[\cdot]$ is the indicator function. A candidate pair $(p,q)$ is considered discordant if its relative ordering is reversed between the two decision vectors. The distance between two individuals is then obtained by aggregating the distances across all decision situations:
{
\setlength{\abovedisplayskip}{4pt}
\setlength{\belowdisplayskip}{4pt}
\begin{equation}
    D_{\tau}(i,j)=\sum_{s=1}^{S}{d_{\tau}(z_{i}^{(s)}, z_{j}^{(s)})}
\end{equation}
}

where $S$ is the number of decision situations, and $z^{(s)}_i$ and $z^{(s)}_j$ denote the decision vectors of individuals $i$ and $j$ in situation $s$, respectively.

\textbf{Pearson and Spearman correlation coefficients}~\cite{pearson1895regression,spearman1904association} are used to measure similarity between decision vectors in priority-value-based and rank-based encodings, respectively. Pearson correlation captures linear relationships between priority values, while Spearman correlation measures monotonic relationships based on rank ordering.
To convert similarity into distance, a unified transformation and aggregation scheme is adopted:
{
\setlength{\abovedisplayskip}{4pt}
\setlength{\belowdisplayskip}{4pt}
\begin{equation}
   D_{\rho}(i,j)=\frac{1}{S}\sum_{s=1}^{S} \frac{1-\rho(z_i^{(s)},z_j^{(s)})}{2}
\end{equation}
where $\rho(\cdot,\cdot)$ denotes either the Pearson or Spearman correlation coefficient, depending on the encoding scheme.
}

\begin{table}
    \caption{Distance metrics for the encoding schemes.}
    \label{tab:distance_functions_encoding}
    \centering
    \begin{tabular}{ll}
    \toprule
    \textbf{Encoding Scheme} & \textbf{Distance Metric} \\
    \midrule
    Priority value     &   Euclidean, Pearson\\
    Rank               &   Euclidean, Kendall tau, Spearman \\
    Binary             &   Euclidean \\
    \bottomrule
    \end{tabular}
\end{table}

\section{Experimental Design}

\subsection{Design of Comparisons}

 This study aims to investigate the impact of different phenotypic characterisation (PC) encoding schemes and distance metrics on surrogate performance. The GP system without surrogate assistance, denoted as GP, is used as the baseline algorithm. The proposed surrogate-assisted GP framework is referred to as SGP. To clearly distinguish different variants, we adopt the naming convention SGP-⟨encoding⟩-⟨distance metric⟩. Based on the combinations considered in this work, six SGP variants are evaluated:
\begin{itemize}
    \item SGP-Value-Euclidean/Pearson
    \item SGP-Rank-Euclidean/Kendall/Spearman
    \item SGP-Binary-Euclidean
\end{itemize}
Furthermore, to examine the effect of surrogate filtering strength, we vary the number of intermediate offspring generated in each generation.  $\lambda_{\text{off}}\in\{2,4\}$ is considered to analyse how different levels of candidate expansion influence the performance of SGP variants.

\subsection{Simulation Model}

To evaluate the performance of GP individuals, the evolved rules are embedded into a discrete-event simulation framework to construct project schedules. The simulation model is capable of loading different project instances and dynamically reproducing their execution processes under uncertainty.
The characteristics of the project instances are mainly determined by three factors: the complexity of precedence relations among activities, activity modes, and resource constraints. The precedence structures are generated using RanGen \cite{demeulemeester_rangen_2003}, a widely used RCPSP instance generator. The complexity of these precedence relations is controlled by the Order Strength (OS) \cite{demeulemeester_rangen_2003} parameter, where a higher OS value indicates denser precedence constraints, while a lower OS allows more activities to be executed concurrently. Following the design principles of the static multi-mode RCPSP benchmark dataset MMLIB \cite{peteghem_experimental_2014}, three levels of precedence complexity are considered, with OS values set to 0.75, 0.5, and 0.25. Each project instance consists of 200 activities.
Regarding activity modes, the expected execution time of the fastest mode is uniformly sampled from the interval [5, 10]. Slower modes have longer durations in exchange for reduced resource consumption. To model uncertainty, the actual duration of each activity is randomly realised within a deviation range of ±3 units from its expected value.
Each project involves 12 types of renewable resources. For each activity, the demand for each resource type is randomly generated within the range [1, 6]. The overall level of resource scarcity is controlled by the Resource Strength (RS) \cite{demeulemeester_rangen_2003} parameter, which is set to 0.25 in this study.

Based on the different levels of precedence complexity, three testing scenarios are defined, denoted as 0.75/R12, 0.5/R12, and 0.25/R12, respectively.
To obtain reliable performance estimates, each GP individual is evaluated on five independent project instances within each scenario. The fitness value is defined as the average makespan normalised by the corresponding lower bound across these instances.
To improve the generalisation ability of the proposed approach, a seed rotation mechanism \cite{hildebrandtImprovedDispatchingRules2010} is adopted. Specifically, during training, the actual durations of activities are resampled using different random seeds at each generation when evaluating GP individuals.
During testing, the same project instances are used; however, each instance is evaluated under 10 independent realisations of activity durations. The average scheduling performance across these realisations is then taken as the objective value on the test instances.

\subsection{Parameter Settings}
The parameters of the GP algorithm are set following commonly adopted configurations in the GP literature \cite{tian_scalable_2025}. The details are shown in Table \ref{tab:GP_parameter}. The features of the project are used as the terminal set of GP. These features are categorised into three groups, namely time-based, precedence-based, and resource-based attributes. Detailed descriptions of these features are provided in Table \ref{tab:terminals}.

To construct phenotypic characterisation (PC) vectors, a set of representative decision situations is first collected. Specifically, for each scenario, project instances are simulated using the LFT and Duration terminals as the activity ordering and activity group selection rules, respectively. All encountered decision situations during these simulations are recorded.
To ensure sufficient discriminatory information, only decision situations with more than 10 candidate items are retained. From the filtered pool, 10 decision situations are randomly selected for each decision type.
These selected decision situations are then fixed and used consistently to compute the PC vectors for all individuals throughout the evolutionary process.

\begin{table}[t]
\caption{The parameter settings of GP.}
\label{tab:GP_parameter}
    \centering
    % \small
    \renewcommand{\arraystretch}{0.85}
    \begin{tabular}{ll}
    \toprule
    \textbf{Parameter}                   & {\textbf{Value}} \\
    \midrule
    Population size                      & 1000 \\
    Number of generations                & 100 \\
    Method for initialising population \quad\quad   & ramped-half-and-half \\
    Initial minimum/maximum depth        & 2 / 6 \\
    Elitism                              & 10 \\
    Maximal program depth                & 8 \\
    Crossover rate                       & 0.80 \\
    Mutation rate                        & 0.15 \\
    Reproduction rate                    & 0.05 \\
    Parent selection                     & Tournament selection (size 7)   \\
    Function set                         & +, -, *, / (protected division)\\
    & \texttt{abs}, \texttt{neg}, \texttt{min} and \texttt{max}\\
    Offspring generation multiplier $\lambda_{\text{off}}$             & 2 or 4 \\
    Number of situations per type        & 10 \\
    \bottomrule
\end{tabular}
\end{table}

\begin{table}[tb]
\caption{Terminal set.}
\label{tab:terminals}
\centering
\begin{tabularx}{0.9\columnwidth}{lll}
\toprule
\textbf{Category} & \textbf{Notation} & \textbf{Description} \\
\midrule
\multirow{7}{*}{Time}
    & EST & Earliest Start Time \\
    & EFT & Earliest Finish Time \\
    & LST & Latest Start Time \\
    & LFT & Latest Finish Time \\
    & Duration   & Expected Duration \\
    & MinDuration & Optimistic Duration \\
    & MaxDuration & Pessimistic Duration \\
    \midrule
\multirow{6}{*}{Precedence}
    & GRPW  & Greatest Rank Positional Weight \\
    & GRPW*  & Greatest Rank Positional Weight All \\
    & TPC  & Total Predecessor Count \\
    & DPC  & Direct Predecessor Count \\
    & TSC  & Total Successor Count \\
    & DSC  & Direct Successor Count \\
    \midrule
\multirow{11}{*}{Resource}
    & RR   & Types of Resource Required  \\
    & GRD   & Greatest Resource Demand  \\
    & AvgRR & Average Resource Requirement \\
    & MaxRR & Maximum Resource Requirement \\
    & MinRR & Minimum Resource Requirement \\
    & AvgRA & Average Resource Availability \\
    & MaxRA & Maximum Resource Availability \\
    & MinRA & Minimum Resource Availability \\
    & AvgRL & Average Resource Left Afterwards \\
    & MaxRL & Maximum Resource Left Afterwards \\
    & MinRL & Minimum Resource Left Afterwards \\
\bottomrule
\end{tabularx}
\end{table}

\section{Experimental Results}

\subsection{Test Performance}
% Binary
% >> Value/rank-euclidean~Rank-spearman
% >> Value Pearson ~ Rank Kendall

\begin{table*}[]
\caption{\textbf{Mean (standard deviation) of the objective values on the test set} obtained by GP and surrogate-assisted GP variants from 30 independent runs.}
\begin{tabular}{llllllllllll}
\hline
\multicolumn{1}{c}{$\lambda_{\text{off}}$} & \multicolumn{1}{c}{Scenario}        & \multicolumn{1}{c}{GP} &  & \multicolumn{2}{c}{Value}                                   &  & \multicolumn{3}{c}{Rank}                                                                                                                                &                      & \multicolumn{1}{c}{Binary}                                               \\ \cline{5-6} \cline{8-10} \cline{12-12} 
\multicolumn{1}{c}{}                       & \multicolumn{1}{c}{}                & \multicolumn{1}{c}{}   &  & \multicolumn{1}{c}{Pearson} & \multicolumn{1}{c}{Euclidean} &  & \multicolumn{1}{c}{Kendall}          & \multicolumn{1}{c}{Spearman}                     & \multicolumn{1}{c}{Euclidean}                                 &                      & \multicolumn{1}{c}{Euclidean}                                            \\ \hline
\multicolumn{1}{c}{2}                      & \textless{}0.75/R12\textgreater{} & 1.724±0.013            &  & 1.716±0.015                 & 1.710±0.012                   &  & 1.708±0.009                          & 1.710±0.010                                      & 1.711±0.014                                                   &                      & 1.704±0.011                                                              \\
                                           &                                     &                        &  & ($\uparrow$)                & ($\uparrow$)($\approx$)       &  & ($\uparrow$)($\uparrow$)($\approx$)  & ($\uparrow$)($\uparrow$)($\approx$)($\approx$)   & ($\uparrow$)($\approx$)($\approx$)($\approx$)($\approx$)      &                      & ($\uparrow$)($\uparrow$)($\uparrow$)($\approx$)($\approx$)($\uparrow$)   \\
                                           & \textless{}0.5/R12\textgreater{}  & 1.691±0.011            &  & 1.682±0.017                 & 1.680±0.016                   &  & 1.677±0.016                          & 1.676±0.015                                      & 1.679±0.017                                                   &                      & 1.671±0.015                                                              \\
                                           &                                     &                        &  & ($\uparrow$)                & ($\uparrow$)($\approx$)       &  & ($\uparrow$)($\approx$)($\approx$)   & ($\uparrow$)($\approx$)($\approx$)($\approx$)    & ($\uparrow$)($\approx$)($\approx$)($\approx$)($\approx$)      &                      & ($\uparrow$)($\uparrow$)($\uparrow$)($\approx$)($\uparrow$)($\uparrow$)  \\
                                           & \textless{}0.25/R12\textgreater{} & 1.710±0.016            &  & 1.712±0.019                 & 1.703±0.014                   &  & 1.699±0.015                          & 1.699±0.014                                      & 1.708±0.013                                                   &                      & 1.692±0.014                                                              \\
                                           &                                     &                        &  & ($\approx$)                 & ($\uparrow$)($\uparrow$)      &  & ($\uparrow$)($\uparrow$)($\uparrow$) & ($\uparrow$)($\uparrow$)($\approx$)($\approx$)   & ($\approx$)($\approx$)($\approx$)($\downarrow$)($\downarrow$) &                      & ($\uparrow$)($\uparrow$)($\uparrow$)($\uparrow$)($\uparrow$)($\uparrow$) \\
                                           & Average Rank                        & 5.42                   &  & 4.74                        & 3.91                          &  & 3.52                                 & 3.48                                             & 4.15                                                          &                      & 2.76                                                                     \\ \hline
4                                          & \textless{}0.75/R12\textgreater{} & 1.724±0.013            &  & 1.724±0.017                 & 1.717±0.016                   &  & 1.711±0.010                          & 1.706±0.011                                      & 1.711±0.010                                                   &                      & 1.699±0.010                                                              \\
                                           &                                     &                        &  & ($\approx$)                 & ($\approx$)($\approx$)        &  & ($\uparrow$)($\uparrow$)($\approx$)  & ($\uparrow$)($\uparrow$)($\uparrow$)($\uparrow$) & ($\uparrow$)($\uparrow$)($\approx$)($\approx$)($\downarrow$)  &                      & ($\uparrow$)($\uparrow$)($\uparrow$)($\uparrow$)($\uparrow$)($\uparrow$) \\
                                           & \textless{}0.5/R12\textgreater{}  & 1.691±0.011            &  & 1.689±0.023                 & 1.679±0.013                   &  & 1.684±0.019                          & 1.673±0.014                                      & 1.678±0.012                                                   &                      & 1.669±0.009                                                              \\
                                           &                                     &                        &  & ($\approx$)                 & ($\uparrow$)($\approx$)       &  & ($\approx$)($\approx$)($\approx$)    & ($\uparrow$)($\uparrow$)($\uparrow$)($\uparrow$) & ($\uparrow$)($\uparrow$)($\approx$)($\approx$)($\approx$)     &                      & ($\uparrow$)($\uparrow$)($\uparrow$)($\uparrow$)($\approx$)($\uparrow$)  \\
                                           & \textless{}0.25/R12\textgreater{} & 1.710±0.016            &  & 1.712±0.025                 & 1.701±0.016                   &  & 1.693±0.015                          & 1.697±0.014                                      & 1.702±0.015                                                   &                      & 1.691±0.013                                                              \\
                                           &                                     &                        &  & ($\approx$)                 & ($\uparrow$)($\approx$)       &  & ($\uparrow$)($\uparrow$)($\uparrow$) & ($\uparrow$)($\uparrow$)($\approx$)($\approx$)   & ($\uparrow$)($\approx$)($\approx$)($\downarrow$)($\approx$)   &                      & ($\uparrow$)($\uparrow$)($\uparrow$)($\approx$)($\approx$)($\uparrow$)   \\
                                           & Average Rank                        & 5.48                   &  & 5.13                        & 4.14                          &  & 3.89                                 & 3.16                                             & 3.83                                                          & \multicolumn{1}{c}{} & 2.36                                                                     \\ \hline
\end{tabular}

\label{tab:SGP_performance_comparison}
\end{table*}

Table \ref{tab:SGP_performance_comparison} presents the test performance of GP and the surrogate-assisted GP variants under two offspring multiplier settings, i.e., $\lambda_{\text{off}}=2 \text{ and } 4$. Each algorithm was independently run 30 times, and the table reports the mean and standard deviation of the average objective values on the test instances. The Wilcoxon signed-rank test was used to compare the algorithms pair-wise. For each algorithm, the statistical test was conducted against the algorithms listed in the preceding columns of the table, where $(\uparrow)$, $(\downarrow)$ and $(\approx)$ indicate that the corresponding algorithm is significantly better, significantly worse or statistically similar to the compared algorithm, respectively. The average rank over all test scenarios is also reported at the bottom of each $\lambda_{\text{off}}$ setting.

Among the surrogate-assisted variants, Binary–Euclidean shows the strongest and most consistent performance. It obtains the lowest mean objective value in all three scenarios for both offspring generation multipliers and achieves the best overall average rank, decreasing from 2.76 with $\lambda_{\text{off}}=2$ to 2.36 with $\lambda_{\text{off}}=4$. The rank-based representations also perform competitively, particularly Rank–Spearman, which achieves the second-best average rank for both multiplier settings. In contrast, the priority-value-based variants are generally less competitive, with Value–Pearson exhibiting only limited improvement over GP in several cases.

Increasing the offspring generation multiplier $\lambda_{\text{off}}$ from 2 to 4 does not lead to a uniform improvement across all surrogate configurations. Binary–Euclidean and Rank–Spearman benefit from the larger intermediate offspring pool, whereas several other combinations show relatively small changes in their final objective values and average ranks. This indicates that simply providing more intermediate offspring for surrogate-based preselection does not necessarily translate into a proportional improvement in optimisation performance.

\begin{figure}[!t]
    \centering
    \includegraphics[width=\linewidth]{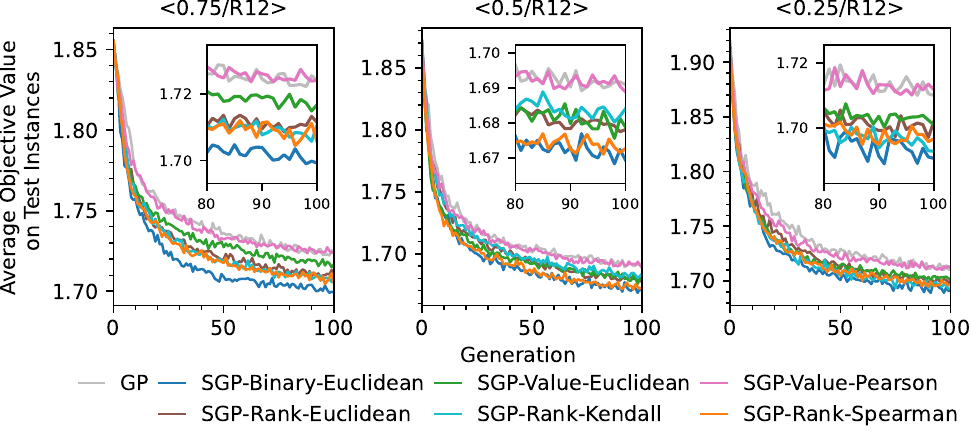}
    \caption{Convergence curves of algorithms with offspring generation multiplier value of 4 over 30 independent runs.}
    \label{fig:SGP_convergence_curve}
\end{figure}

Figure \ref{fig:SGP_convergence_curve} further illustrates the convergence behaviour of the algorithms with $\lambda_{\text{off}}=4$. The SGP variants generally achieve lower objective values than GP at the same generation, corresponding to the same budget of full fitness evaluations. Binary–Euclidean consistently maintains one of the lowest objective values from the early-to-middle stages of evolution and preserves this advantage until the end of the run across all three scenarios. This observation is consistent with its superior final objective values and average rank in Table \ref{tab:SGP_performance_comparison}.
The rank-based variants also show favourable convergence behaviour, particularly Rank–Spearman, whereas Value–Pearson remains close to standard GP for a substantial portion of the evolutionary process.

The above results demonstrate that the effectiveness of surrogate-assisted GP depends strongly on the encoding–distance combination. However, the final optimisation performance alone does not reveal why some surrogate configurations are consistently more effective than others, or why increasing the number of intermediate offspring does not always yield a corresponding improvement. We therefore conduct further analyses of the surrogate models and the resulting offspring populations to investigate the mechanisms underlying these differences.

\subsection{Population Improvement}

\begin{figure}
    \centering
    \includegraphics[width=\linewidth]{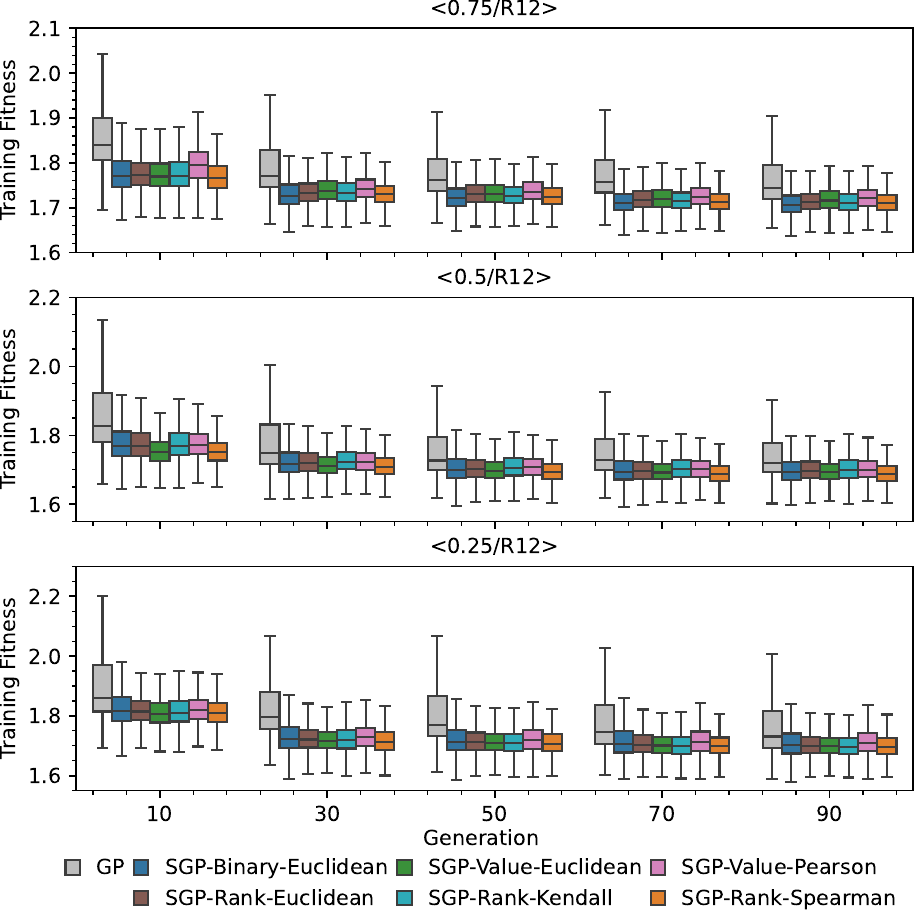}
    \caption{Training fitness distribution of the GP and the surrogate-assisted GP variants when $\lambda_{\text{off}}=4$ across generations over 30 independent runs.}
    \label{fig:SGP_population_improvement}
\end{figure}

The previous convergence analysis mainly focuses on the test performance of the best individual found in each generation. To further examine whether the surrogate-assisted GP improves the overall population quality, Figure~\ref{fig:SGP_population_improvement} compares population training-fitness distributions at generations 10, 30, 50, 70, and 90 over 30 runs with $\lambda_{\text{off}}=4$. Because seed rotation changes the training environment, comparisons are restricted to algorithms within the same generation. The SGP variants generally have lower fitness distributions than GP, indicating improvements in overall population quality as well as in the best individuals.
This suggests that the surrogate-assisted mechanism not only improves the quality of the best individual, as shown in the convergence curves, but also raises the overall quality of the evolving population.

\subsection{Computational Overhead}

\begin{table*}
    \caption{\textbf{Mean runtime} of real evaluation, PC vector generation and surrogate estimation per generation in seconds.}
    \setlength{\tabcolsep}{5.8pt}
    \begin{tabular}{lllllllllllll}
\hline
$\lambda_{\text{off}}$  & Scenario             & \multicolumn{3}{c}{<0.75/R12>}      &  & \multicolumn{3}{c}{<0.5/R12>}       &  & \multicolumn{3}{c}{<0.25/R12>}      \\ \cline{3-5} \cline{7-9} \cline{11-13} 
 & Algorithm            & Real Eval. & PC Gen. & Surr. Est. &  & Real Eval. & PC Gen. & Surr. Est. &  & Real Eval. & PC Gen. & Surr. Est. \\ \cline{1-5} \cline{7-9} \cline{11-13} 
2 & SGP-Binary-Euclidean & 401.22     & 11.93   & 0.24       &  & 819.37     & 33.31   & 0.28       &  & 936.53     & 15.14   & 0.25       \\
  & SGP-Rank-Kendall     & 378.40     & 14.31   & 1.96       &  & 754.60     & 28.85   & 22.85      &  & 1162.31    & 15.92   & 4.07       \\
  & SGP-Rank-Spearman    & 349.65     & 12.83   & 4.20       &  & 634.82     & 28.30   & 8.83       &  & 938.37     & 15.65   & 5.70       \\
  & SGP-Rank-Euclidean   & 411.17     & 14.12   & 0.30       &  & 790.67     & 34.84   & 0.32       &  & 1153.11    & 18.94   & 0.32       \\
  & SGP-Value-Pearson    & 342.78     & 12.26   & 0.74       &  & 715.75     & 27.13   & 0.90       &  & 1012.03    & 16.07   & 0.92       \\
  & SGP-Value-Euclidean  & 301.69     & 14.40   & 0.31       &  & 728.33     & 35.59   & 0.36       &  & 956.74     & 19.87   & 0.33       \\
4 & SGP-Binary-Euclidean & 348.20     & 16.24   & 0.41       &  & 655.57     & 52.37   & 0.44       &  & 854.87     & 25.09   & 0.42       \\
  & SGP-Rank-Kendall     & 361.38     & 17.79   & 3.20       &  & 629.27     & 50.86   & 43.09      &  & 1015.25    & 24.30   & 7.61       \\
  & SGP-Rank-Spearman    & 380.73     & 19.09   & 7.76       &  & 617.20     & 51.90   & 14.92      &  & 1097.80    & 24.92   & 11.01      \\
  & SGP-Rank-Euclidean   & 403.78     & 22.57   & 0.52       &  & 808.14     & 63.50   & 0.55       &  & 1257.30    & 30.49   & 0.52       \\
  & SGP-Value-Pearson    & 374.23     & 19.53   & 1.52       &  & 823.69     & 51.54   & 1.75       &  & 1003.12    & 28.68   & 1.90       \\
  & SGP-Value-Euclidean  & 355.74     & 19.17   & 0.46       &  & 614.12     & 65.14   & 0.64       &  & 969.94     & 26.20   & 0.48       \\ \hline
\end{tabular}

    \label{tab:overhead_running_time}
\end{table*}

Table \ref{tab:overhead_running_time} reports the average runtime per generation for real fitness evaluation, PC vector generation, and surrogate estimation. Overall, the additional computational cost introduced by SGP is small compared with real fitness evaluation. Across all configurations, the combined runtime of PC vector generation and surrogate estimation remains substantially lower than that of real evaluation. Duplicate removal is not separately reported because it is implemented using vectorised NumPy \cite{harris2020array} operations and takes approximately one second per generation on average.
PC vector generation accounts for most of the additional overhead and requires a similar amount of time across different SGP variants. Its computational cost increases when $\lambda_{\text{off}}$ increases from 2 to 4, as more intermediate offspring need to be characterised. Nevertheless, PC generation remains considerably less expensive than real fitness evaluation.

In contrast, surrogate estimation time is more dependent on the distance metric. Euclidean and Pearson distances are computationally inexpensive, whereas the rank-based Spearman and Kendall distances require substantially more time. This is because Spearman distance requires ranking operations, with approximately $O(\sum_{s=1}^{S} n_s\log n_s)$ complexity, while the implemented Kendall distance involves pairwise comparisons with approximately $O(\sum_{s=1}^{S} n_s^2)$ complexity. Consequently, Rank-Kendall is particularly expensive when the PC vectors contain situations with large candidate sets.

\section{Further Analysis}
\subsection{Intermediate Offspring Analysis Setup}
The main experimental results show clear performance differences among the SGP variants, but do not explain how the choice of PC encoding and distance measure leads to these differences. We therefore conduct further analyses to examine two main mechanisms of the proposed framework: the ability of different encoding--distance combinations to estimate offspring fitness for preselection, and the effect of different PC encodings on duplicate removal and offspring diversity.
To isolate these effects from differences in evolutionary trajectories, the analyses of surrogate fitness estimation, duplicate removal, and offspring diversity are conducted on the same sets of intermediate offspring generated from baseline GP populations. Specifically, the populations from all 30 independent GP runs are collected every 10 generations, and 4000 intermediate offspring are generated from each sampled population. For the 2000-offspring setting, the first 2000 offspring from each set are used as a subset. The true fitness values of all 4000 offspring are evaluated, allowing different PC encodings and distance measures to be applied retrospectively to exactly the same offspring sets at each pool size. The results reported in the following analyses are aggregated over the 30 independent runs.

\subsection{Surrogate Fitness Estimation and Preselection Quality}

\begin{figure}
    \centering
    \includegraphics[width=\linewidth]{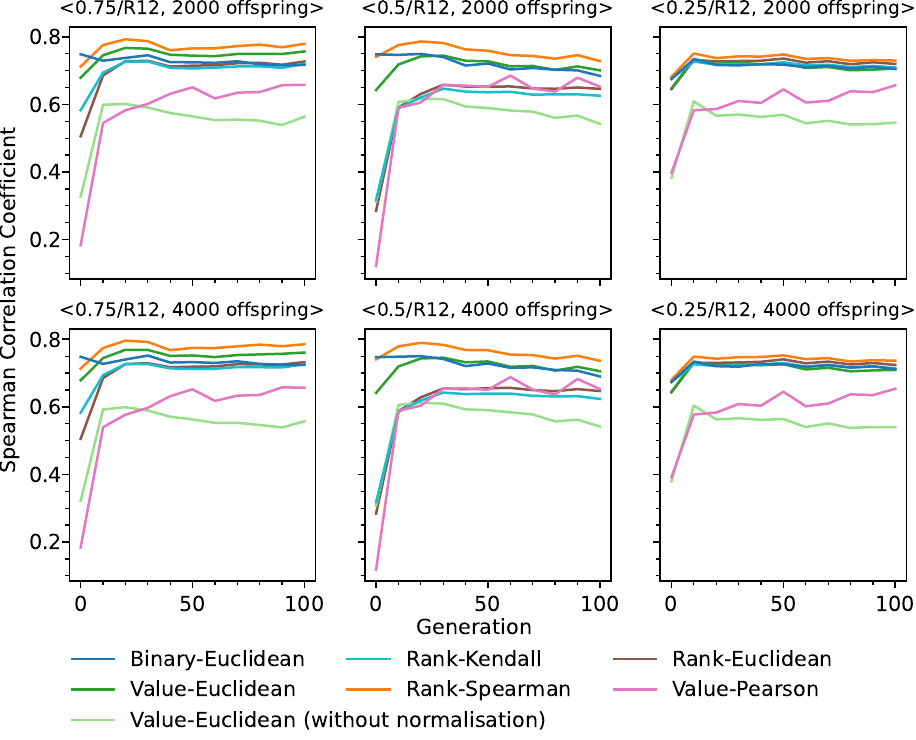}
    \caption{Mean Spearman correlation between true and estimated offspring fitness across generations over 30 independent runs.}
    \label{fig:SGP_spearman_analysis}
\end{figure}

We first examine how the choice of PC encoding and distance measure affects surrogate fitness estimation. Figure~\ref{fig:SGP_spearman_analysis} shows that Rank--Spearman generally achieves the highest correlations across the three scenarios and both offspring-pool sizes, particularly after the initial generation. Normalised Value--Euclidean and Binary--Euclidean also maintain relatively high correlations throughout evolution. Rank--Euclidean and Rank--Kendall show more scenario-dependent performance, becoming competitive in the <0.25/R12> scenario, where the differences among these five combinations are small. Value--Pearson exhibits weak correlations initially, although its ranking accuracy improves over subsequent generations. Similar patterns are observed for both offspring-pool sizes.

For the priority-value encoding, normalised Value--Euclidean consistently outperforms its unnormalised counterpart in Figure~\ref{fig:SGP_spearman_analysis}. This supports applying situation-wise normalisation before concatenating the decision vectors into the complete PC vector. Without normalisation, decision situations with larger priority-value ranges may dominate the Euclidean distance, weakening the correspondence between behavioural similarity and fitness similarity.

% Scatter plot of a specific run at gen 0
\begin{figure}
    \centering
    \includegraphics[width=\linewidth]{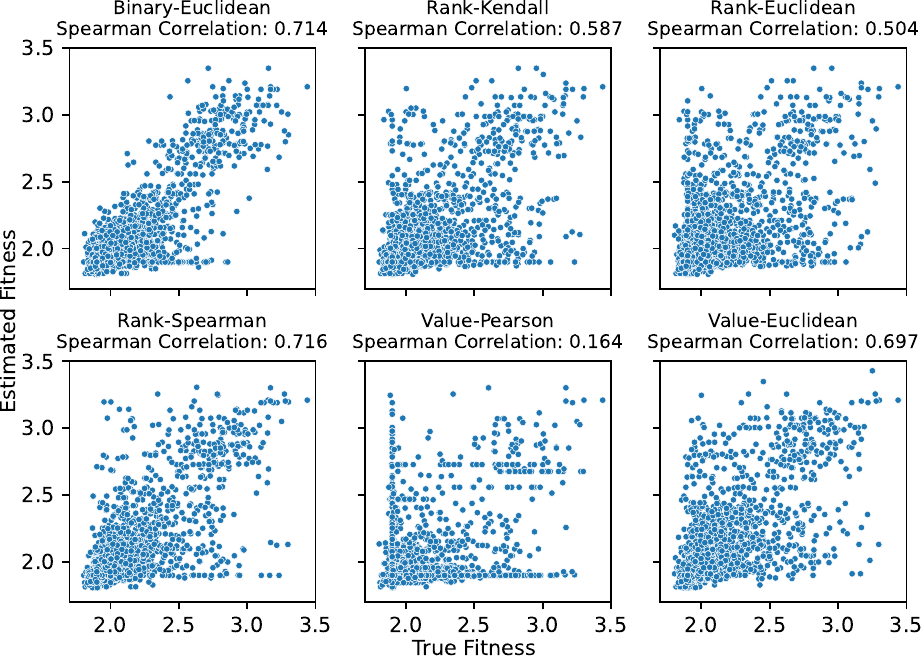}
    \caption{An illustrative example of true versus estimated fitness from one sampled offspring set at Generation 0 for <0.75/R12>.}
    \label{fig:SGP_estimation_scatter}
\end{figure}

Figure~\ref{fig:SGP_estimation_scatter} illustrates these estimation behaviours using one sampled offspring set at Generation 0 in the <0.75/R12> scenario. Rank--Spearman, Binary--Euclidean, and Value--Euclidean exhibit clearer positive associations between true and estimated fitness than Value--Pearson, consistent with the early-generation patterns in Figure~\ref{fig:SGP_spearman_analysis}. In particular, Value--Pearson assigns similar estimates to offspring with substantially different true fitness values, limiting its ability to distinguish their relative quality.

% Top 1000 overlap
\begin{figure}
    \centering
    \includegraphics[width=\linewidth]{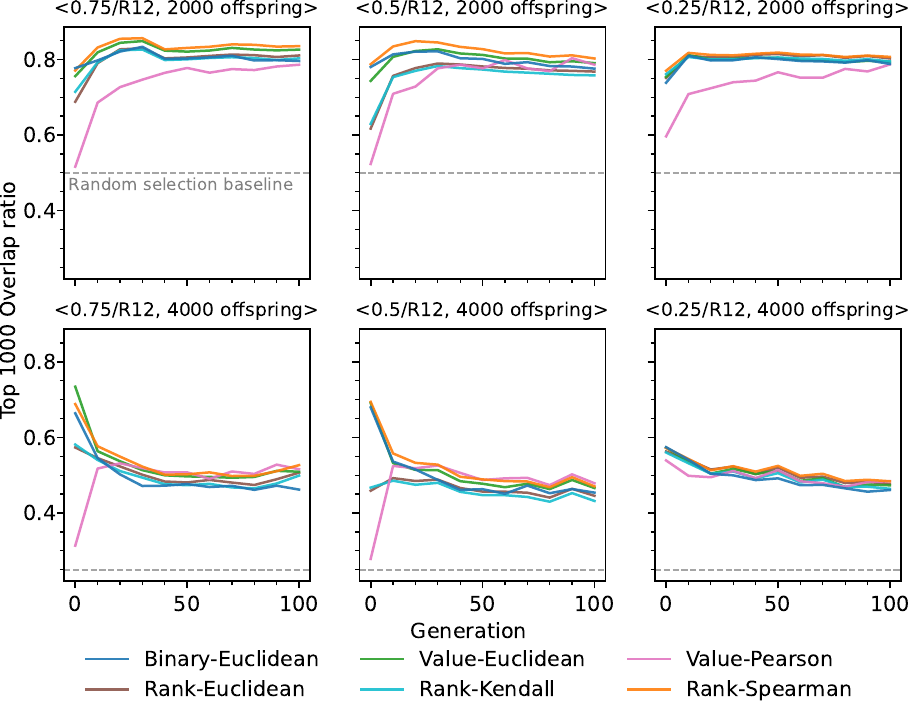}
    \caption{Ratio of true top-1000 offspring identified by different encoding-distance combinations across generations.}
    \label{fig:top_1000_overlap_ratio}
\end{figure}

We next examine whether stronger global ranking accuracy translates into better offspring preselection. Figure~\ref{fig:top_1000_overlap_ratio} reports the proportion of true top-1000 offspring retained by surrogate preselection. With 2000 intermediate offspring, Rank--Spearman generally achieves the highest overlap ratios, while Value--Euclidean and Binary--Euclidean also perform well. Most variants attain ratios of approximately 0.75--0.85 after the early generations, above the random-selection baseline of 0.5. With 4000 intermediate offspring, the random-selection baseline is 25\%. The variants achieve similar overlap ratios of approximately 0.45--0.55 in later generations, all above this baseline.

These results distinguish global fitness-ranking accuracy from preselection quality. Global correlation measures the ordering of the entire offspring pool, whereas preselection depends on which offspring belong to the selected top subset. Consequently, higher global correlation need not yield proportionally better top-1000 retention. Furthermore, Rank--Spearman's stronger global ranking accuracy does not translate into the best overall test performance, which is achieved by Binary--Euclidean in Table~\ref{tab:SGP_performance_comparison}. Surrogate estimation accuracy alone therefore does not fully explain the performance differences among SGP variants, motivating the following analyses of duplicate removal and behavioural diversity.

\subsection{Duplicate Removal Capacity}

\begin{figure}
    \centering
    \includegraphics[width=\linewidth]{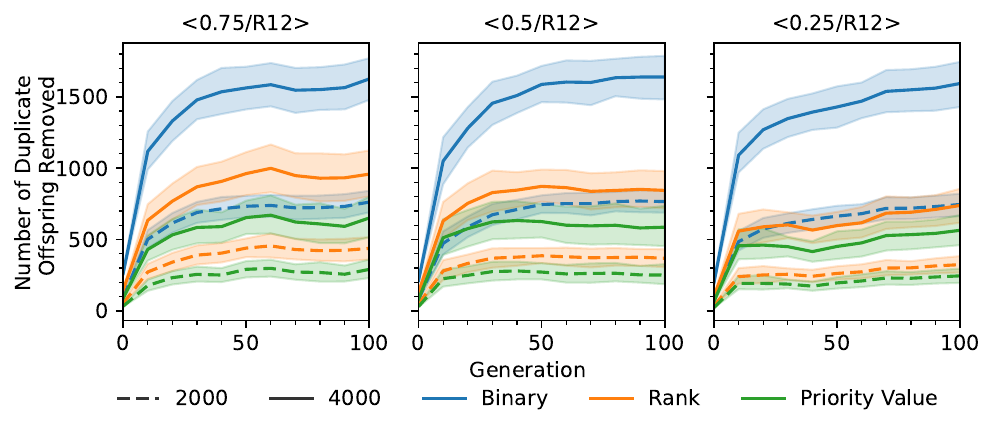}
    \caption{Number of duplicate intermediate offspring removed across generations.}
    \label{fig:duplicate_removal_counts}
\end{figure}

In addition to surrogate fitness estimation, the PC encoding also determines which intermediate offspring are regarded as phenotypically identical during duplicate removal. This may affect the composition of the candidate pool before preselection. We therefore examine the duplicate removal capacity of the three PC encoding schemes using the same intermediate offspring sets described previously.
Figure~\ref{fig:duplicate_removal_counts} shows clear and consistent differences among the encoding schemes. Binary encoding removes the largest number of duplicate offspring, followed by rank encoding, while priority-value encoding removes the fewest. This ordering is observed across all three scenarios and for both 2000 and 4000 intermediate offspring. As expected, generating 4000 offspring results in substantially more duplicates than generating 2000, with the increase being particularly pronounced for binary encoding. The number of duplicates also generally increases over generations, indicating that more intermediate offspring become behaviourally similar as evolution progresses.

These differences arise from the level of behavioural detail retained by each encoding. Binary encoding records only the final selected items in the sampled decision situations, so GP individuals with different priority values or rankings can still produce identical binary PC vectors. Rank encoding preserves more information about the relative ordering of candidates, reducing the likelihood of identical representations. Priority-value encoding retains the exact priority values and is therefore the most fine-grained representation, making exact duplicate PC vectors less frequent. Consequently, the choice of PC encoding substantially affects how much redundancy is removed from the intermediate offspring pool before surrogate-based preselection.

\subsection{Offspring Diversity}
\begin{figure}
    \centering
    \includegraphics[width=\linewidth]{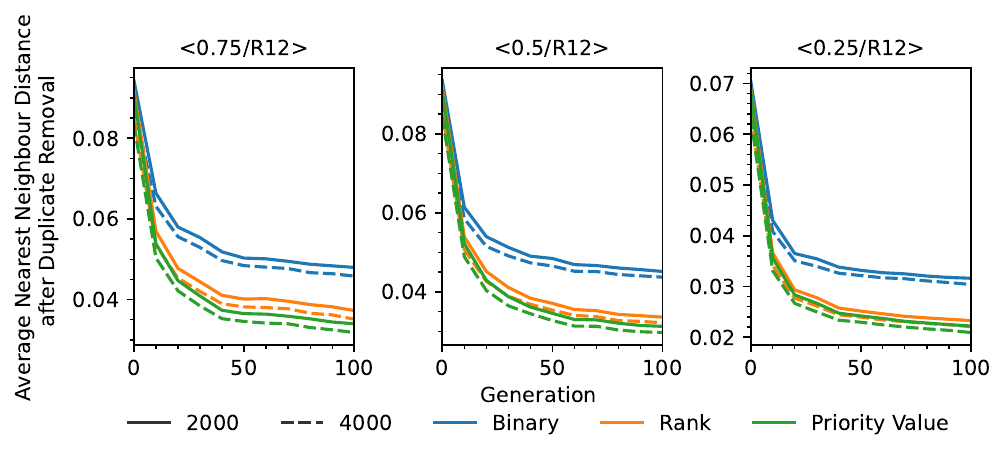}
    \caption{Normalised average nearest distance between intermediate offspring after duplicate removal.}
    \label{fig:average_nearest_distance_after_duplicate_removal}
\end{figure}

\begin{figure}
    \centering
    \includegraphics[width=\linewidth]{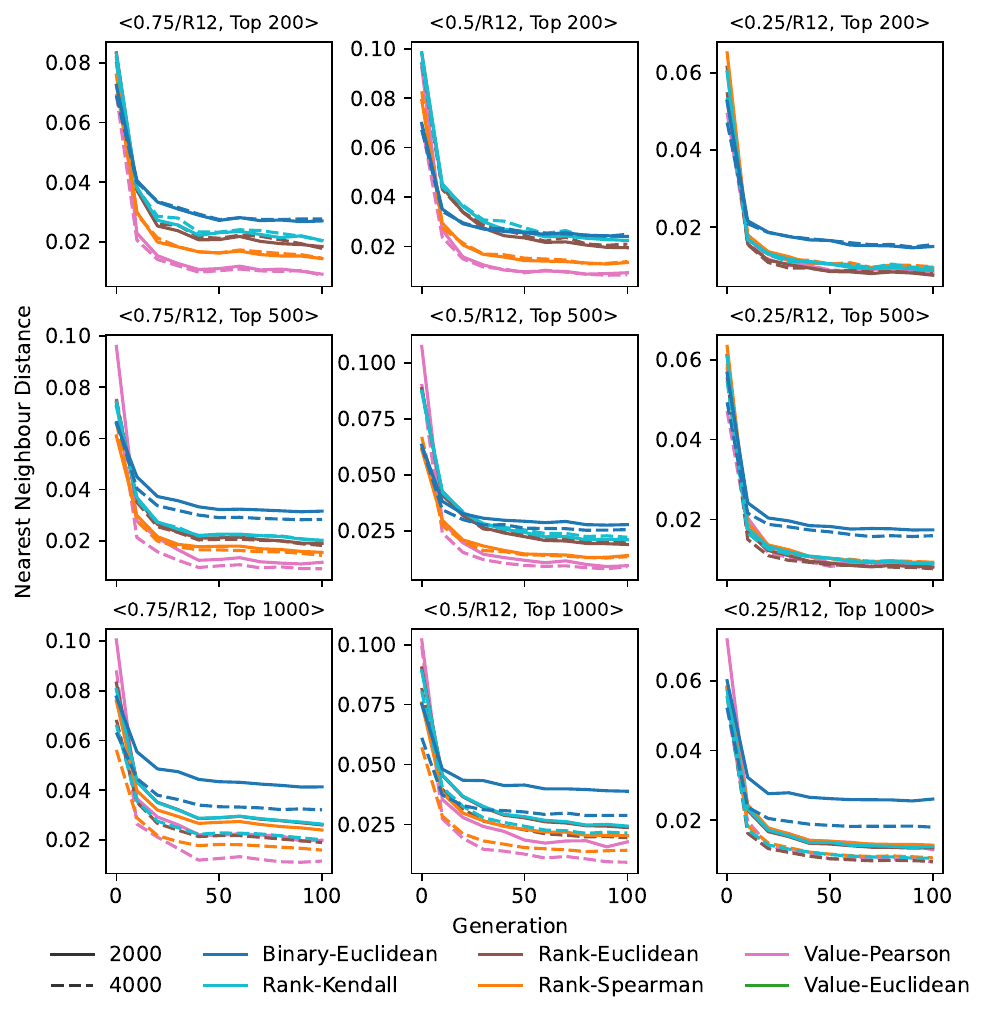}
    \caption{Normalised average nearest distance between intermediate offspring after duplicate removal and surrogate preselection.}
    \label{fig:average_nearest_distance_after_duplicate_removal_surrogate_preselection}
\end{figure}

We next examine whether the different duplicate-removal behaviours observed in Figure~\ref{fig:duplicate_removal_counts} lead to differences in offspring diversity. To make the diversity values comparable across different encoding--distance combinations, all retained offspring are converted to the binary PC representation and their distances are measured using Euclidean distance. We use the normalised average nearest-neighbour distance to measure local behavioural diversity, where a larger value indicates that the retained offspring are less crowded in the behavioural space.
Figure~\ref{fig:average_nearest_distance_after_duplicate_removal} shows the diversity of the intermediate offspring after duplicate removal. Clear differences can be observed among the three encoding schemes. Binary encoding consistently yields the largest average nearest-neighbour distance, followed by rank encoding, while priority-value encoding gives the smallest values. This pattern is consistent across all three scenarios and for both 2000 and 4000 intermediate offspring. These results show that the stronger duplicate-removal capacity of binary encoding indeed reduces local redundancy more effectively and leaves a more behaviourally diverse intermediate-offspring pool. The average nearest-neighbour distance also decreases over generations for all three encodings, indicating that the offspring become increasingly concentrated in similar behavioural regions as evolution progresses.

Figure~\ref{fig:average_nearest_distance_after_duplicate_removal_surrogate_preselection} further examines the diversity of the offspring retained after both duplicate removal and surrogate preselection. The analysis is conducted on the true top-200, top-500, and top-1000 offspring within the selected set. These three subset sizes are used to examine the degree of crowding at different quality levels. In particular, the top-200 and top-500 subsets are relevant because tournament selection with size 7 is expected to draw most parents for subsequent reproduction from the top-500 region. Across all scenarios, the average nearest-neighbour distance is generally smallest for the top-200 subset, followed by the top-500 and top-1000 subsets. This suggests greater local behavioural crowding among higher-quality offspring in the sampled sets under the common Binary--Euclidean representation.

Despite the increasing crowding among higher-quality offspring, Binary--Euclidean generally maintains larger average nearest-neighbour distances after preselection, particularly in later generations. Its diversity advantage is evident not only in the top-1000 subset but also in the top-500 subset across the three scenarios. For the top-200 subset, Binary--Euclidean also preserves greater diversity in the <0.75/R12> and <0.25/R12> scenarios, while Binary--Euclidean, Rank--Kendall, and Rank--Euclidean exhibit similar diversity in the <0.5/R12> scenario. These results show that its diversity advantage extends to the high-quality offspring most likely to be selected as parents under tournament selection. Thus, binary-based duplicate removal combined with surrogate preselection preserves behavioural diversity in the regions particularly relevant to subsequent evolution. Another notable observation is that the average nearest-neighbour distance is often smaller when $\lambda_{\text{off}}=4$ than when $\lambda_{\text{off}}=2$, even after preselection. This suggests that generating more intermediate offspring does not necessarily yield a more diverse set of promising retained offspring, which may help explain why increasing $\lambda_{\text{off}}$ does not always produce proportionally better optimisation performance.

\subsection{Ablation Study}

To empirically examine the contributions of duplicate removal and surrogate preselection to the performance of the proposed SGP framework, we conduct an ablation study using different combinations of these components. Except for the baseline GP, all variants use the binary PC encoding with Euclidean distance and $\lambda_{\text{off}}=4$. In addition, a perfect-preselection variant is included, in which the true fitness values of all intermediate offspring are used for preselection instead of surrogate estimates.

\begin{table}[]
    \caption{\textbf{Mean (standard deviation) objective values on the test set} of the ablation variants over 30 independent runs. Binary PC encoding, Euclidean distance and $\lambda_{\text{off}}=4$ are used.}
    \centering
    {%
\setlength{\tabcolsep}{3pt}% horizontal padding inside cells
\renewcommand{\arraystretch}{0.95}% vertical padding between rows
\begin{tabular}{@{}llll@{}}
\hline
                                             & \multicolumn{3}{c}{Scenario}                                                                                                                        \\ \cline{2-4} 
Algorithm                                    & \textless{}0.75/R12\textgreater{}             & \textless{}0.5/R12\textgreater{}              & \textless{}0.25/R12\textgreater{}             \\ \hline
GP                                           & 1.724±0.013                                     & 1.691±0.011                                     & 1.710±0.016                                     \\
GP + Duplicate Removal                         & 1.715±0.011                                     & 1.683±0.012                                     & 1.705±0.015                                     \\
                                             & ($\uparrow$)                                    & ($\uparrow$)                                    & ($\approx$)                                     \\
GP + Surrogate Preselection                              & 1.722±0.014                                     & 1.687±0.012                                     & 1.715±0.020                                     \\
                                             & ($\approx$)($\downarrow$)                       & ($\approx$)($\approx$)                          & ($\approx$)($\downarrow$)                       \\
GP + Duplicate Removal                         & 1.699±0.010                                     & 1.669±0.009                                     & 1.691±0.013                                     \\
 \quad + Surrogate Preselection & ($\uparrow$)($\uparrow$)($\uparrow$)            & ($\uparrow$)($\uparrow$)($\uparrow$)            & ($\uparrow$)($\uparrow$)($\uparrow$)            \\
GP + Duplicate Removal                       & 1.698±0.014                                     & 1.666±0.014                                     & 1.685±0.013                                     \\
\quad + Perfect Preselection                        & ($\uparrow$)($\uparrow$)($\uparrow$)($\approx$) & ($\uparrow$)($\uparrow$)($\uparrow$)($\approx$) & ($\uparrow$)($\uparrow$)($\uparrow$)($\approx$) \\ \hline
\end{tabular}
}

    \label{tab:ablation_study_comparison}
\end{table}

\begin{figure}
    \centering
    \includegraphics[width=\linewidth]{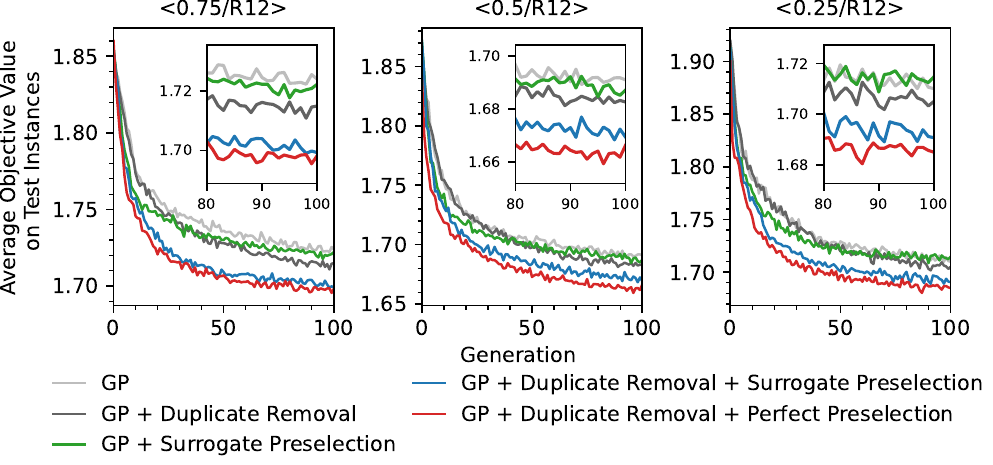}
    \caption{Convergence curves for the ablation study of duplicate removal and preselection with $\lambda_{\text{off}}=4$ over 30 independent runs.}
    \label{fig:ablation_study_convergence_curve}
\end{figure}

Table~\ref{tab:ablation_study_comparison} and Figure~\ref{fig:ablation_study_convergence_curve} show that duplicate removal alone provides a more consistent improvement over standard GP than surrogate preselection alone. GP + Duplicate Removal significantly outperforms GP in the <0.75/R12> and <0.5/R12> scenarios and achieves comparable performance in <0.25/R12>. In contrast, GP + Surrogate Preselection remains statistically comparable to GP across the three scenarios. It is also worse than GP + Duplicate Removal in two scenarios. These results indicate that, under the examined configuration, removing phenotypically redundant offspring makes a stronger individual contribution to search performance than applying surrogate preselection without duplicate removal.

More importantly, combining duplicate removal with surrogate preselection produces substantially better results than using either component independently. GP + Duplicate Removal + Surrogate Preselection significantly outperforms GP, GP + Duplicate Removal, and GP + Surrogate Preselection in all three scenarios. This demonstrates that the two mechanisms are complementary: duplicate removal reduces behavioural redundancy in the intermediate offspring pool, while surrogate preselection directs the remaining evaluation budget towards promising offspring. Surrogate preselection therefore becomes substantially more effective when operating on the less redundant candidate pool produced by duplicate removal.

The perfect-preselection variant further evaluates whether the remaining surrogate estimation error limits optimisation performance. It achieves slightly lower mean objective values than surrogate-based preselection in all three scenarios, and its convergence curves in Figure~\ref{fig:ablation_study_convergence_curve} generally decrease faster and remain slightly below those of the surrogate-based variant. However, the differences in final objective values are not statistically significant. This suggests that more accurate fitness estimation can still improve the convergence trajectory, but the current Binary--Euclidean surrogate already provides sufficiently effective preselection such that replacing its estimates with true fitness values yields only limited additional improvement.

\subsection{Discussion}
The further analyses show that the performance differences among the SGP variants cannot be explained by surrogate fitness estimation accuracy alone. Although the encoding--distance combinations exhibit noticeable differences in Spearman correlation, their ability to retain the true top offspring becomes considerably more similar after the early generations. Moreover, replacing surrogate estimates with true fitness values in the perfect-preselection variant produces only a limited additional improvement in final solution quality. These results suggest that the current surrogate model is already sufficiently accurate for guiding the evolutionary search, and further improvements in its ranking accuracy are unlikely to yield substantial additional benefits.

A second important finding is that the PC encoding affects not only surrogate estimation but also how phenotypic redundancy is defined and removed. Binary encoding removes substantially more duplicate offspring than rank and priority-value encodings. When the retained offspring are evaluated in a common Binary--Euclidean behavioural space, stronger duplicate removal is associated with larger nearest-neighbour distances, indicating lower local redundancy. This effect remains visible after surrogate preselection, including among the high-quality offspring that are most likely to contribute to subsequent generations. The ablation study further supports this interpretation: duplicate removal alone provides a more consistent improvement than surrogate preselection alone, while their combination achieves the strongest performance. Therefore, duplicate removal and surrogate preselection appear to play complementary roles, with the former reducing behavioural redundancy and the latter directing the evaluation budget towards promising candidates.

These findings also help explain why increasing $\lambda_{\text{off}}$ does not always lead to a proportional improvement in optimisation performance. A larger intermediate-offspring pool provides more candidates for preselection, but also contains more duplicates and closely related offspring. After preselection, the retained offspring with $\lambda_{\text{off}}=4$ are often no more diverse, and sometimes less diverse, than those obtained with $\lambda_{\text{off}}=2$. Therefore, increasing the number of generated candidates is beneficial only when the framework can effectively control redundancy and preserve useful diversity among promising offspring.

From a practical perspective, PC encodings and distance measures should therefore be selected according to their effect on the complete surrogate-assisted evolutionary process rather than surrogate accuracy alone. Among the configurations investigated in this study, Binary--Euclidean provides the most favourable overall trade-off. It achieves consistently strong optimisation performance and competitive preselection quality, while the binary representation enables substantially stronger duplicate removal and better preservation of behavioural diversity. Euclidean distance also incurs considerably lower computational overhead than the rank-based correlation distances. Binary--Euclidean is therefore recommended as the default configuration for the proposed SGP framework, while the extent to which this advantage generalises to other GP domains and behavioural characterisations remains an important direction for future investigation.

\section{Conclusions}
This study developed a phenotypic-characterisation-based surrogate-assisted GP framework for dynamic multi-mode project scheduling, combining duplicate removal with 1-NN-based offspring preselection. Across the examined scenarios, Binary--Euclidean achieved the strongest and most consistent optimisation performance while introducing relatively low computational overhead. Further analyses indicate that this advantage cannot be explained by surrogate prediction accuracy alone: binary encoding removed more phenotypically redundant offspring and preserved greater behavioural diversity. The ablation results further showed that duplicate removal and surrogate preselection are complementary, with their combination producing the largest improvement. These findings suggest that PC representations should be evaluated by their effects on the overall evolutionary search rather than by surrogate prediction accuracy alone.

Future work will first investigate adaptive PC construction, in which decision situations are updated during evolution and weighted or replaced according to their discriminative ability. Second, adaptive redundancy and diversity management will be explored by adjusting the intermediate-offspring size according to population diversity and incorporating both surrogate-predicted fitness and behavioural novelty into offspring preselection. Finally, noise-aware surrogate models will be developed to explicitly account for the stochastic fitness variation introduced by seed rotation.

\bibliographystyle{ieeetr}
\bibliography{references}

\end{document}